# A new design of a fall detection system integrating landmark identification and deep learning techniques

**Tri Nhut Do[1*], Thi Thuy Le[2]**

[1]*University of Information Technology, Vietnam National University in Ho Chi Minh City, Quarter 34, Linh Xuan Ward, Ho Chi Minh City, Vietnam*
[2]*Thu Dau Mot University, 6 Tran Van On Street, Phu Loi Ward, Ho Chi Minh City, Vietnam*

Received 12 December 2024; revised 11 April 2025; accepted 23 April 2025

***Abstract:***

**This article introduces an innovative system that integrates landmark identification with deep learning to enhance fall detection accuracy and reliability. By utilising advanced computer vision techniques, such as MediaPipe for spatial recognition, the system effectively differentiates between routine movements and actual falls. The integration of landmarks with a deep learning prediction algorithm minimises false alarms, ensuring timely responses to genuine falls. Comprehensive experimentation underscores the system's versatility across various scenarios, emphasising its potential to improve safety and independence for older adults. The training process demonstrates a steady increase in accuracy, stabilising by the 40th cycle, while error rates decline significantly during the initial cycles. Real-time experiments, involving both male and female participants aged 8 to 50, recorded a remarkable 95% detection rate of falls, showcasing the system's effectiveness and promising future applications in elder care and smart health monitoring environments.**

## 1. Introduction

The global ageing population, projected to reach 1.5 billion by 2050, faces a growing public health crisis from falls, which the World Health Organisation (WHO) identifies as a leading cause of injury-related deaths, claiming 646,000 lives annually, predominantly among the elderly [1]. In Vietnam, 1.5-1.9 million older adults experience falls annually, with 5% requiring hospitalisation, and at the University Medical Centre of University of Medicine and Pharmacy at Ho Chi Minh city, 17% of monthly geriatric admissions are fall-related [2, 3]. These alarming statistics underscore the urgent need for robust fall detection systems (FDS) to enhance safety, reduce response times, and promote independent living for older adults.

Current FDS, primarily wearable devices and ambient sensors, suffer from significant limitations. Wearable systems, such as accelerometers, are hindered by user discomfort, limited battery life (12-24 hours), and high false-positive rates (up to 20%), as they struggle to distinguish falls from routine activities [4]. Ambient sensors, using pressure mats or infrared, offer non-intrusive monitoring but are prone to accuracy issues in dynamic indoor settings, with false-positive rates reaching 15% due to environmental variability [5]. Recent advancements in computer vision and deep learning have shown promise in overcoming these shortcomings. K. Chouhan, et al. (2022) [4] demonstrated real-time fall detection using image recognition, achieving 85% accuracy in controlled settings but struggling with real-world variability. C. Ge, et al. (2018) [5] applied convolutional neural networks (CNNs) to differentiate falls from non-falls, reporting 90% precision in laboratory environments; however, their models faltered in untrained scenarios due to limited contextual data. Pose estimation techniques, such as those using MediaPipe, have furthered progress by identifying key body points to track movement, but standalone implementations often lack integration with predictive models, resulting in delayed or inaccurate responses. These studies highlight a critical gap: the need for systems that combine contextual awareness, real-time processing, and robust generalisation to diverse environments.

This study addresses these gaps and challenges by proposing a novel vision-based FDS that integrates landmark identification with deep learning to enhance fall detection for elder care. The main objectives are: (1) To develop a non-invasive system using MediaPipe's pose estimation and CNN-based prediction to accurately distinguish falls from routine movements; (2) To minimise false positives through contextual analysis of spatial body keypoints; (3) To achieve real-time detection with low latency (<500 ms) for timely interventions; and (4) To enable scalability for IoT-integrated smart home deployment. The primary contributions include: (1) A hybrid framework combining pose estimation and deep learning for superior fall detection accuracy; (2) A context-aware approach reducing false alarms compared to wearable (20% false positives) and ambient (15% false positives) systems; (3) Real-time performance optimised for diverse indoor settings, surpassing lab-constrained prior work [6]; and (4) A scalable design compatible with IoT ecosystems, facilitating seamless monitoring in elder care settings. Unlike existing systems, our approach is non-invasive, user-friendly, and adaptable, fostering greater safety and independence for the elderly while setting a new benchmark for vision-based fall detection. Furthermore, its potential integration with IoT frameworks [7-9] enables seamless monitoring, paving the way for scalable deployment in smart homes and healthcare facilities. This research not only advances fall detection technology but also fosters greater independence and safety for the elderly, aligning with global health priorities.

*Corresponding author: Email: trinhutdo@uit.edu.vn, trinhutdo@gmail.com

This article is structured as follows: Section 2 provides a comprehensive background review of fall detection technologies, including an analysis of existing systems. Section 3 details the design of the proposed fall detection system, including both hardware and software components. Section 4 presents experimental results that demonstrate the system's accuracy and robustness in pose estimation. Finally, Section 5 concludes the research, outlining future directions for improving the system and expanding its application in smart health monitoring environments. Through this work, we aim to fill the current gaps in fall detection technology by offering an innovative, accurate, and user-friendly solution to enhance elderly care and safety.

## 2. Background review

Falls are a leading cause of injury-related deaths and disabilities, particularly among the elderly. As the global population of older adults continues to increase, the prevalence of falls also rises, making fall detection systems a critical component of public health. The WHO reports that annually, approximately 646,000 people die of falls, with older adults being the most affected group [10]. In addition to fatalities, falls lead to serious injuries such as fractures and head trauma, which can significantly reduce the quality of life and independence for elderly individuals [11].

Traditional fall detection systems typically fall into two categories: wearable devices and ambient systems. Wearable devices, including accelerometers and gyroscopes, monitor real-time movement patterns. These devices have shown effectiveness in detecting falls but face limitations, such as user discomfort, limited battery life, and difficulties in distinguishing falls from routine movements, often resulting in false alarms [12]. Additionally, wearable devices require users to wear uncomfortable sensors, posing a barrier to widespread adoption, especially among older adults.

Ambient systems, which rely on environmental sensors like cameras, pressure mats, and infrared sensors, offer a less intrusive approach. However, these systems often struggle with accuracy due to fluctuating environmental factors such as lighting conditions or obstacles [13]. Furthermore, they are prone to false positives, where routine actions like sitting or bending down are mistakenly classified as falls, triggering unnecessary interventions [14].

In recent years, computer vision has emerged as a promising alternative for fall detection, particularly using MediaPipe, a machine learning framework developed by Google. MediaPipe provides advanced techniques for real-time pose estimation and body tracking, which are essential for accurately identifying and tracking human movements. Its Pose Estimation model tracks 33 body keypoints in real time, offering precise insights into an individual's posture and movements [15]. This makes MediaPipe a valuable tool for detecting sudden changes in posture or body movement, key indicators of falls.

Several studies have demonstrated the effectiveness of MediaPipe's Pose Estimation model in fall detection. For example, P. Sirikongtham, et al. (2025) [16] showed how MediaPipe's ability to track body points could help identify falls by recognising significant changes in body posture. This approach has a significant advantage in reducing false positives, as it can distinguish between normal activities and genuine falls more accurately than traditional sensor-based systems. MediaPipe's integration with deep learning models, such as Long Short-term Memory (LSTM) networks, further improves the system's ability to process movement data over time, enhancing prediction accuracy for falls.

MediaPipe offers several advantages over conventional sensor-based systems. It is a cost-effective solution because it uses standard cameras for monitoring and reducing the need for expensive specialised sensors. Moreover, cameras are easier to install and do not require users to wear any devices, making the system more acceptable, especially for elderly individuals who may find wearable devices uncomfortable. Additionally, computer vision-based systems like MediaPipe provide a wider field of view, which enhances their ability to monitor larger areas and detect falls more comprehensively [17].

Despite the benefits, challenges remain in adopting MediaPipe-based fall detection systems. Lighting conditions and camera placement are crucial for system accuracy. Poor lighting or improper camera positioning can lead to missed detections or false alarms. Moreover, environmental obstructions, such as furniture or other people, can hinder the system's ability to track movements effectively [18]. These limitations, however, can be mitigated through proper calibration and advanced image processing techniques.

To enhance the effectiveness of MediaPipe-based systems, future research should focus on integrating IoT technologies and edge computing. This integration would allow for real-time processing at the network edge, enabling immediate alerts and improving response times [19]. Furthermore, using multi-camera systems could improve accuracy by providing multiple perspectives of the environment, helping to capture all angles of potential falls [20].

Moreover, deep learning techniques should continue to evolve to improve the robustness and accuracy of fall detection systems. The development of advanced models like Generative Adversarial Networks (GANs) for generating synthetic training data can help improve the model's ability to recognise a wider range of fall scenarios, making it more adaptable to different real-world conditions [21].

In conclusion, computer vision, particularly Google's MediaPipe, offers a promising alternative. Its Pose Estimation model tracks 33 body keypoints in real-time, enabling precise fall detection by analysing posture changes. Studies, such as P. Sirikongtham, et al. (2025) [16], demonstrate MediaPipe's ability to reduce false positives through contextual analysis, enhanced by LSTM networks for temporal modelling. Unlike costly sensor-based systems, MediaPipe uses standard cameras, ensuring cost-effectiveness and user comfort. However, challenges like lighting and obstructions require advanced calibration.

This article introduces a novel fall detection system that combines MediaPipe's pose estimation with deep learning algorithms, specifically for landmark identification and movement analysis. This approach offers a non-invasive, highly accurate, and user-friendly solution for fall detection, which addresses the limitations of current systems. By providing real-time insights into user movements and accurately distinguishing between routine activities and falls, this system holds significant promise for enhancing elderly care and health monitoring.

## 3. System design

The Fall Detection System that is shown in Fig. 1 can be classified based on various criteria, including the technology used, the application domain, and the method of detection [22]. By branching of the Fall Detection System, vision-based Fall Detection Systems utilise computer vision techniques to analyse video feeds from cameras to identify falls in real-time.

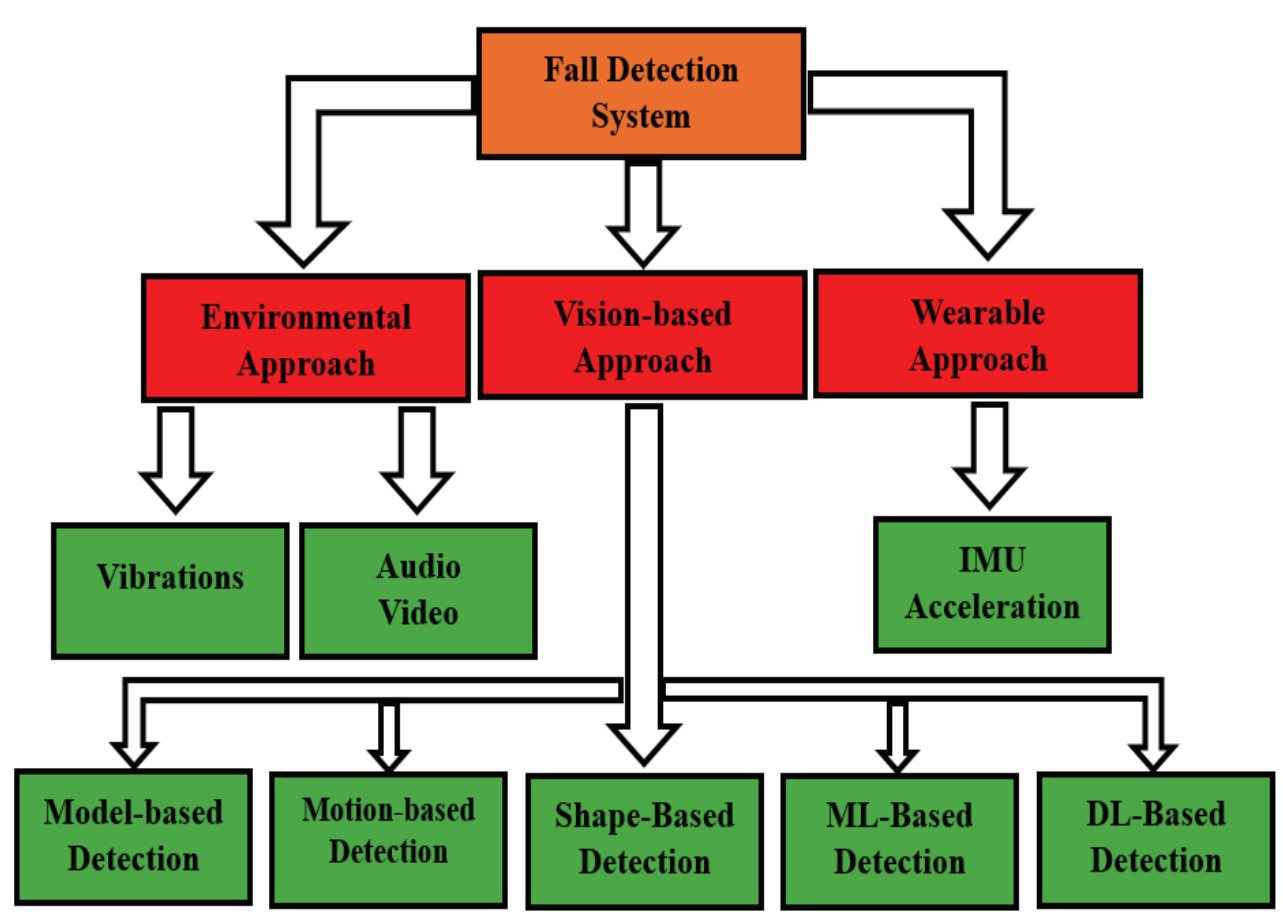

**Fig. 1. Types of Fall Detection Systems.** Source: The authors.

The block diagram illustrated in Fig. 2 represents a Fall Detection System that effectively monitors individuals using video input from a camera. The system is structured sequentially, comprising several key components that work together to identify falls, trigger alarms, and alert users. The system begins with a video or camera input block, where the camera captures real-time footage of the monitored area. This input can come from various types of cameras, including stationary, PTZ (pan-tilt-zoom), or wearable cameras. The purpose of this block is to continuously stream video data, which serves as the primary source for real-time analysis.

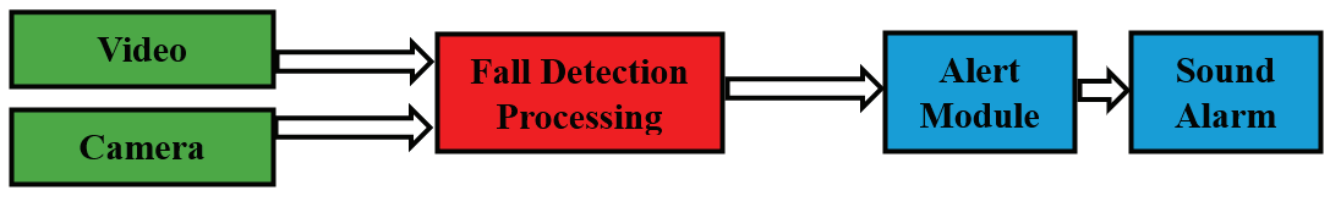

**Fig. 2. Fall Detection System block diagram.** Source: The authors.

The second block is Fall Detection Processing. This block acts as the core processing unit of the system, utilising computer vision and machine learning algorithms to analyse the incoming video feed. The fall detection processing block identifies patterns and movements indicative of a fall (for instance, an abrupt change in the individual's posture, velocity, or position). It processes frames in real time to flag potential falls by comparing the current motion data with predefined models of normal and abnormal movements.

The next block is the alert module. Once a fall is detected, the information is channelled into the alert module, which acts as a decision-making driver. This module can operate based on different criteria, such as the severity of the fall or the time elapsed since detection. Based on the detection outcomes, the alert module determines the appropriate response. It may also evaluate whether to send a notification to caregivers or emergency services, depending on the configuration of the system.

The final block is the sound alarm block, which serves as the auditory output component of the system. It is triggered by the alert module when a fall is confirmed. The sound alarm produces a loud acoustic signal designed to alert nearby individuals or caregivers. This immediate response is crucial in ensuring timely assistance for the person who has fallen.

In conclusion, the block diagram of the Fall Detection System illustrates a streamlined process in which video captured from a camera is analysed in real time to detect falls, prompting an alert that activates a sound alarm. This integrated approach enhances safety and response times in environments where fall incidents can occur, such as homes, hospitals, or elder care facilities.

### *3.1. Fall Detection Processing design*

The proposed system for fall detection leverages advanced computer vision techniques and a deep learning-based LSTM algorithm [23] to facilitate real-time monitoring for elder care, as illustrated in Fig. 3. Utilising a 1080p RGB camera that captures video frames at 30 frames per second, the system employs MediaPipe's Pose Estimation to extract 33 body keypoints (such as shoulders and hips) with over 90% accuracy, providing 3D coordinate data. The sequences of keypoints are then normalised, filtered using methods such as Kalman smoothing, and segmented into one-second windows (approximately 30 frames) for temporal analysis.

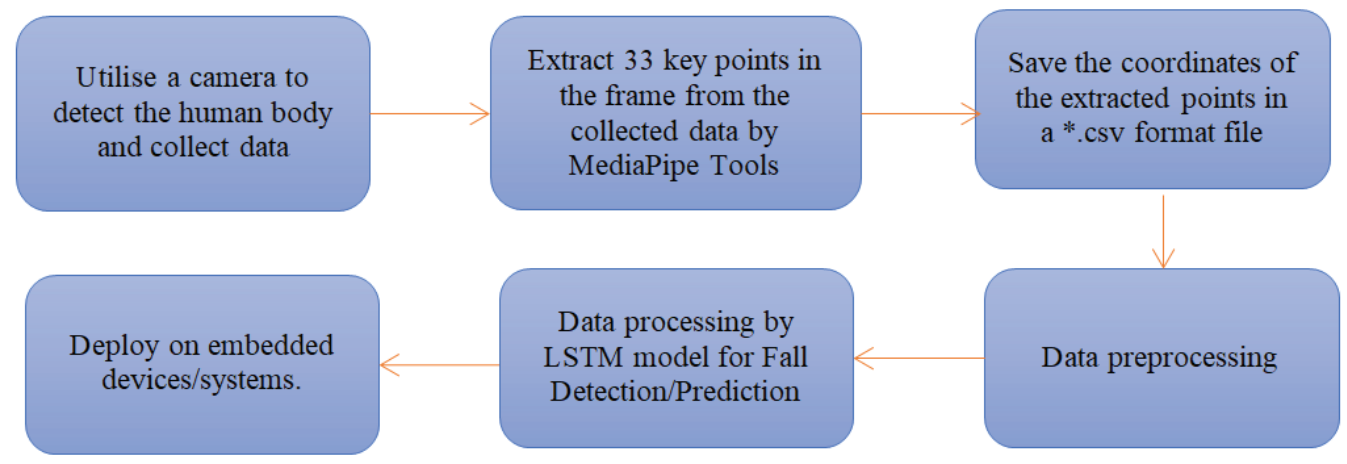

**Fig. 3. Human body motion recognition system diagram with MediaPipe tools and Long Short-term Memory model.** Source: The authors.

This vision-based fall detection system represents a significant leap forwards compared to state-of-the-art (SOTA) solutions in elder care. Unlike traditional wearable systems, which often lead to user discomfort and experience high false-positive rates of up to 20% due to a lack of contextual awareness [24], or ambient sensors, which are affected by environmental variability and result in around 15% false positives [25], our design is both non-invasive and context-aware. In contrast to existing vision-based methods, which provide 85% accuracy but are constrained by real-world variability [4], and C. Ge, et al. (2018) [5]'s system, which achieves 90% precision yet is limited to laboratory conditions, our approach demonstrates superior generalisation across diverse indoor environments through robust preprocessing techniques and adaptive modelling.

### *3.2. MediaPipe Pose*

MediaPipe tools are an open-source, cross-platform machine learning framework developed by Google that focuses on building real-time multimedia applications for mobile devices and the web. It offers a variety of multimedia processing modules and advanced machine learning algorithms, including applications in the fields of facial recognition, pose estimation, and gesture recognition. MediaPipe can handle data from various sources, such as images, videos, audio, and sensor data, facilitating real-time machine learning algorithms. It supports cross-platform computation, is efficient and scalable, and can be deployed on multiple hardware platforms, including mobile devices, desktops, and embedded systems, while also supporting programming languages such as C++, Python, Java, and JavaScript.

One of the important features of MediaPipe tools is MediaPipe Pose, a specialised module for detecting and tracking human body posture. MediaPipe Pose uses deep learning algorithms, specifically CNNs, to identify a total of 33 keypoints on the human body from video or images. These keypoints include significant positions on the body, such as the head, eyes, shoulders, elbows, hips, knees, and ankles. Fig. 4 illustrates these locations.

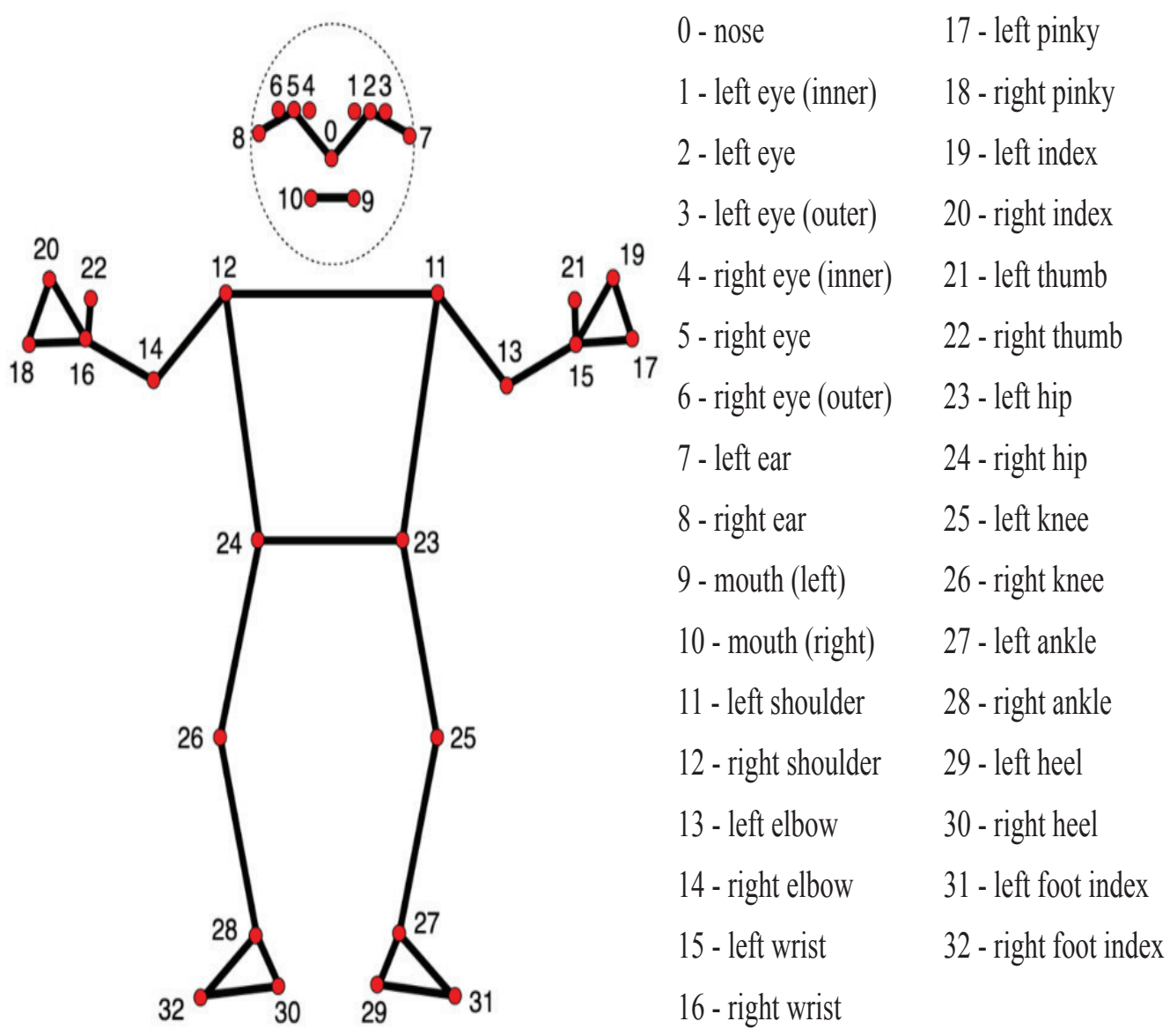

**Fig. 4. Body keypoints based on MediaPipe Pose.** Source: MediaPipe.

These body keypoints are used to estimate the user's posture and analyse their movements in real time. This is important for applications such as fall detection, posture monitoring in sports, and medical assistance in evaluating patient movement. By relying on real-time data from videos, MediaPipe Pose can provide accurate and timely information about body posture.

### *3.3. Modelling - Data capturing and labelling*

To ensure the model can flexibly accommodate a diverse range of users, self-collected data was added to the training dataset. This data includes individuals of varying sizes and genders, encompassing tall and short people, as well as those who are large and slim. We have made efforts to integrate our dataset with the official datasets Le2i and Multiple Cameras Fall Dataset to obtain the coordinates of keypoints for individuals in different locations and postures.

MediaPipe Pose uses CNN to identify and locate 33 key points on the human body, as illustrated in Fig. 4. Each key point is represented by four values: x, y, z, and visibility. With this mapping, each video frame can be transformed into a feature vector comprising 132 features. This mapping process is repeated for all 520 videos, each containing 50 frames and belonging to one of two classes ("fall" and "no fall"). As a result, we have a vector database with a size of 26,000×132.

To standardise the length of the video segments, we selected 50 consecutive frames of the falling action from each video, corresponding to approximately one and a half seconds of duration. This process helps eliminate unnecessary frames at the beginning and end of each video, when the person is standing before the fall and lying on the ground after the fall. This standardisation process is applied to both classes, "fall" and "no fall". The method of extracting 50 frames for a falling action is illustrated in Fig. 5.

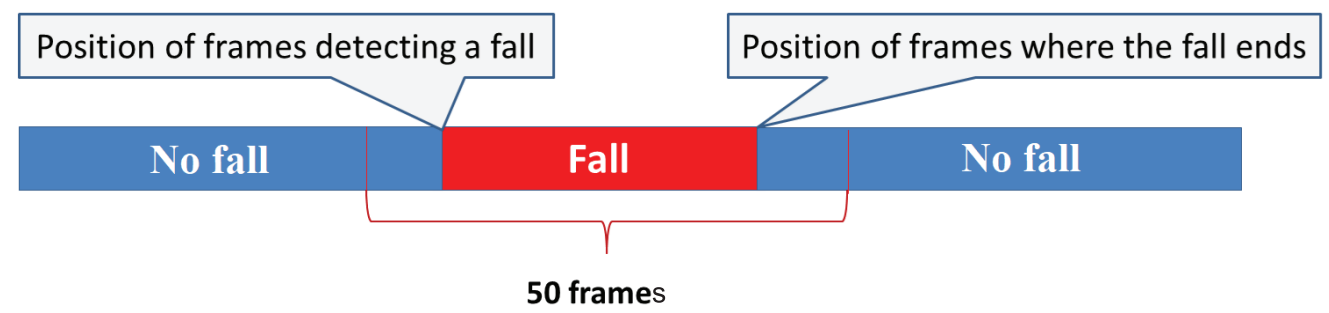

**Fig. 5. The sampling method for fall detection.** Source: The authors.

The overlapping of five to ten frames before and after the fall is intended to enhance the accuracy and recognition capability of the model while ensuring that no transitional states between different phases of the fall are missed.

To prepare the input data for the model, each video, after being standardised into 50 frames, is divided into sequences of 30 frames by shifting ten frames between consecutive sequences, as illustrated in Fig. 6. This method captures the complete states before, during, and after the fall, ensuring data comprehensiveness. Simultaneously, creating overlapping sequences increases the number of samples, thereby improving the model's learning ability and accuracy. With this approach, each 50-frame video can generate three sequences, contributing to the expansion of the training dataset and enhancing the model's performance in fall behaviour recognition.

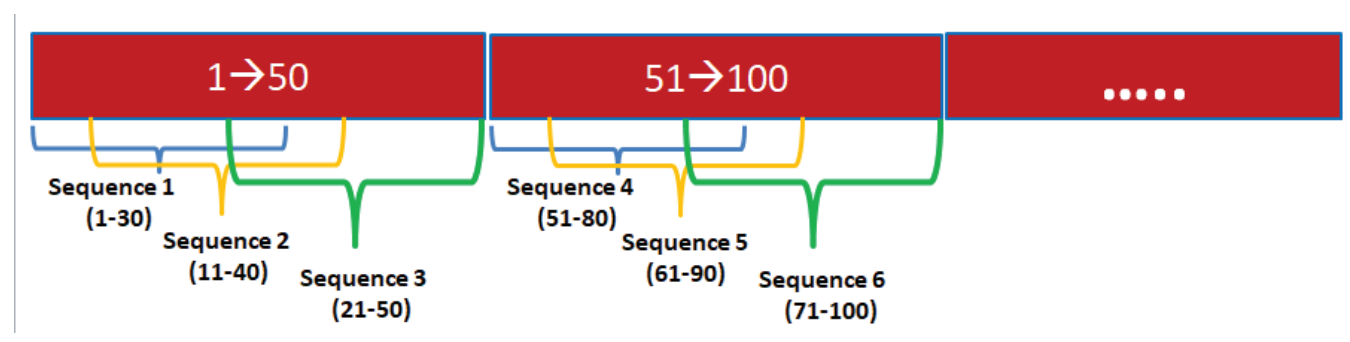

**Fig. 6. Sequence generation from consecutive frames.** Source: The authors.

These sequences are labelled according to the two classes, "fall" and "no fall", forming a dataset with a total of 26,000 samples, each containing 132 features. The application of overlapping and sequence generation techniques not only enriches the dataset but also plays a crucial role in improving the model's learning efficiency, enabling it to operate more accurately in real-world scenarios.

### *3.4. Modelling - Training and testing dataset*

A data array of 33×4 was prepared for each frame to train the model. Here, '33' corresponds to the 33 keypoints on the human body, as shown in Fig. 4, and '4' corresponds to the (x, y, z) coordinates and display confidence. In one training process, we utilised 26,000 data arrays prepared in this manner. The training dataset was divided into two parts, comprising 70% of the dataset for training and 30% for testing. The input training images were preprocessed, and skeletal features were extracted to display indicators of what was predicted to be a fall action. This deep learning model was commonly tested using a test dataset, which included calculated metrics such as True Positives, True Negatives, False Positives, and False Negatives. Additionally, appropriate performance metrics such as Accuracy, F1 Score, Precision, Recall, and Matthews Correlation Coefficient (MCC) were also used to measure the performance of the deep learning model.

### *3.5. Modelling - Long Short-term Memory*

Long Short-term Memory networks were introduced by Hochreiter and Schmidhuber (1997) and are commonly referred to as LSTM, as shown in Fig. 7. This deep learning architecture is considered an extension of Recurrent Neural Networks (RNNs) with the ability to learn long-term dependencies. LSTM is designed to process and remember long-term data sequences, addressing the limitations of traditional RNNs in retaining long-term information.

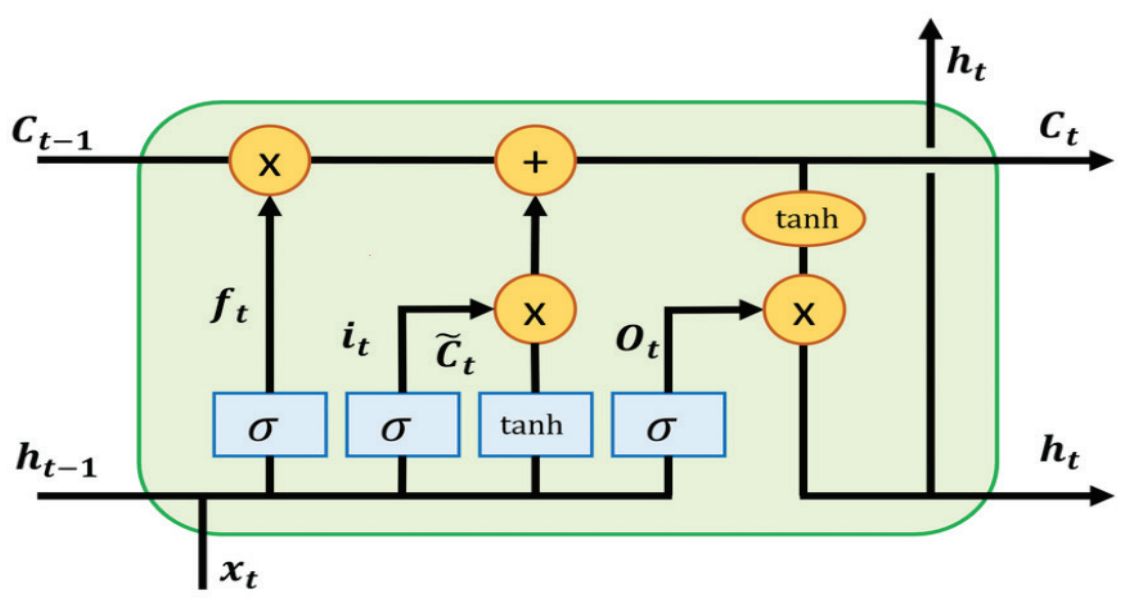

**Fig. 7. A schematic of Long Short-term Memory network - a deep learning architecture.** Source: Geeks for Geeks.

The basic structure of an LSTM model includes input layers, hidden layers, and output layers. The input layer typically consists of input data and time step information. The input data can be sequential, such as text, images, and audio, along with the time step information, which is intended to distinguish the input data at different time steps. The hidden layer comprises multiple LSTM cells, each with a cell state and three gates known as the forget gate, input gate, and output gate. From there, the flow of information is controlled.

*Forget gate:* This gate determines which parts of the previous cell state $c_{t-1}$ should be retained or discarded. The gate operates based on the following formula:

$$f_t = \sigma(W_f \, . \, [h_{t-1}, x_t] + b_f) \quad (1)$$

where σ is the sigmoid function, $W_f$ is the weight; $h_{t-1}$ is the previous hidden state; $x_t$ is the current input at time step t; and $b_f$ is the tuning coefficient of the forget gate.

*Input gate:* This gate monitors new information from the current input $x_t$ and determines what information will be added to the cell state. The formula for calculating the input gate is as follows:

$$i_t = \sigma(W_i \, . \, [h_{t-1}, x_t] + b_i) \quad (2)$$

At the same time, a candidate value for the cell state $\tilde{C}_t$ is generated via the tanh function as follows:

$$\tilde{C}_t = \tanh(W_c \, . \, [h_{t-1}, x_t] + b_c) \quad (3)$$

The current cell state $C_t$ is then updated by combining the value from the forget gate and the input gate, as follows:

$$c_t = f_t * c_{t-1} + i_t * \tilde{C}_t \quad (4)$$

*Output gate:* This gate decides which part of the cell state will be used as the current output $h_t$. The output gate is computed as follows:

$$o_t = \sigma(W_o \, . \, [h_{t-1}, x_t] + b_o) \quad (5)$$

where the current hidden state $h_t$ is calculated by:

$$h_t = o_t * \tanh(c_t) \quad (6)$$

In conclusion, we find that, thanks to these control gates, LSTM can efficiently maintain long-term information, making it a powerful tool for tasks that require processing long data sequences, such as speech recognition, natural language processing, and time series prediction.

## 4. Experimental results and evaluation

Designing experimental scenarios for the computer vision-based fall detection system proposed in this article involves several aspects of computer vision, deep learning, and real-world scenario testing. The following are several experimental scenarios:

The first goal of the experiment in this section was to develop a robust training dataset that could generalise well across various scenarios. Therefore, a diverse dataset was created by capturing video footage of fall and non-fall scenarios in a variety of settings, including homes, gyms, and aged care facilities. It is essential to include variations in lighting, angles, clothing, and body types.

The next objective of the experiment in this section is to evaluate and compare the accuracy, precision, recall, and F1 score of the trained model to determine the most effective architecture for fall detection. As a result, the LSTM deep learning architecture network was deployed and trained for fall detection.

The ultimate goal of the experiments in this section is to examine the differences in fall types by identifying weaknesses in detecting specific fall types and proposing targeted improvements. Therefore, experiments on images and image sequences (videos) were conducted. From this, we analyse the performance of the system in detecting different fall types (backward fall, forward fall, and side fall) by simulating these falls in a controlled environment.

### 4.1. System initial setup

Our experimental environment utilises the Windows 11 operating system, equipped with an 11th-generation Intel(R) Core(TM) i5-1135G7 processor running at 2.40 GHz, 16 GB of RAM, and Intel(R) Iris(R) Xe Graphics.

For this research work, we utilised the HIKVision DS-2CV1021G0-IDW1 camera, which offers a good compromise in terms of maximum resolution of 2 MP (1920×1080 pixels) and is capable of up to 30 frames per second (fps) at full HD resolution. This camera is suitable for a variety of surveillance applications. In our experiment, the camera operates at a frame rate of 25 frames per second (fps) along with a 2 MP resolution. This frame rate is sufficient to analyse smooth motion, allowing for accurate fall detection without any frame lag. The 2 MP resolution provides the detailed imagery necessary for identification tasks, while the 25 fps frame rate contributes to smooth and clear motion capture. Both factors significantly influence the system's overall performance, impacting not only the quality of the surveillance footage but also the demands on storage and processing capabilities.

The programming language used is Python 3.12.3. The LSTM model is built using the TensorFlow framework, employing the LSTM layer from the Keras library. The sigmoid activation function is applied in the gates of the LSTM to control the 'forget' and 'save' operations of information across time steps, enabling the network to selectively learn important information from the input data sequence.

Some of the key parameters of the LSTM network used in the experiment to train the model are given in Table 1.

**Table 1. Key parameters of the Long Short-term Memory network for training.**

| Parameters | Values |
|---|---|
| The number of frames in a data sequence | 30 |
| Batch size | 64 |
| Features | 4 |
| Number of Epochs | 80 |
| Activation function | Sigmoid |
| Model optimisation | Adam |
| Number of LSTM layers | 4 |
| Number of units per LSTM layer | 128 |

### 4.2. Training validation and experimental results evaluation

In this section, the results obtained from the experiments on the test set are statistically presented and represented by the confusion matrix, as shown in Fig. 8. This confusion matrix provides a visual representation of actual versus predicted classifications, facilitating a thorough analysis of the model's accuracy and types of errors. It illustrates the model's performance when classifying between two classes: "fall" and "background" (or no fall).

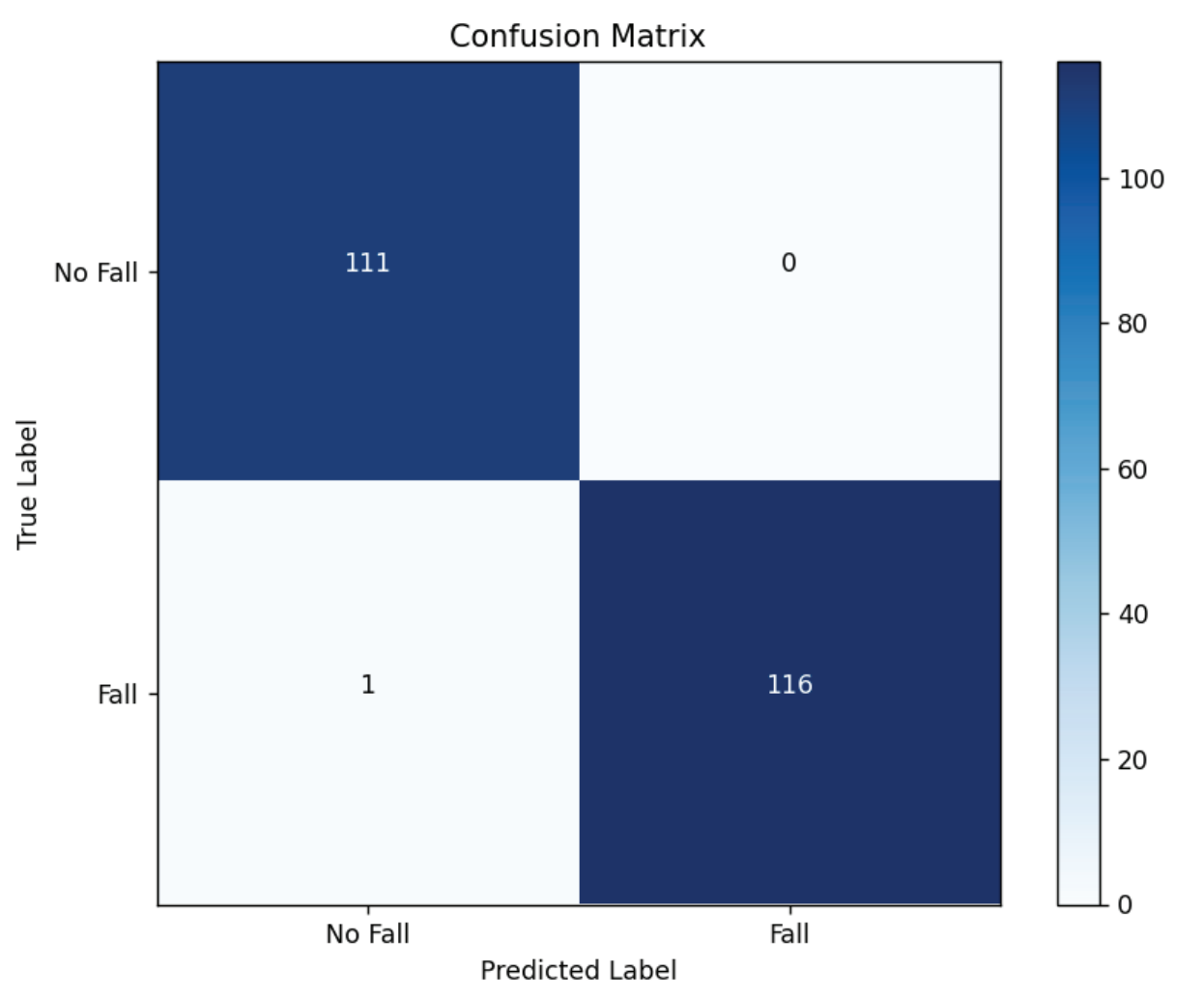

**Fig. 8. Confusion matrix.** The vertical axis (True label) is the actual label; The horizontal axis (Predicted label) is the label predicted by the model; The cells represent the corresponding number of samples; Specific values in the matrix are listed in Table 2. Source: The authors.

**Table 2. Confusion matrix specific values.**

| | True: Fall | True: No-fall |
|---|---|---|
| Predicted: Fall | 116 (True positive) | 0 (False positive) |
| Predicted: No Fall | 1 (False negative) | 111 (True negative) |

Explanation: 116 (TP): The model correctly predicted the image as a fall; 0 (FP): The model mistook an image without a fall as having a fall; 1 (FN): The model failed to detect a fall when one was actually present; 111 (TN): The number of samples that were actually "no fall" but were correctly predicted as "no fall" - the model correctly predicted the image as a no fall.

Evaluation: The model is quite effective at classifying falls and no falls, as evidenced by the high number of true positives and true negatives.

The accuracy and error rates on the training and testing datasets after each training cycle are shown in Figs. 9 and Fig. 10, respectively. The accuracy for both training and testing steadily increases after each training cycle. By the 40th training cycle, the accuracy begins to stabilise. In terms of error rates during training and testing, these significantly decrease during the first ten training cycles. Around the 45th cycle and onwards, they also start to stabilise.

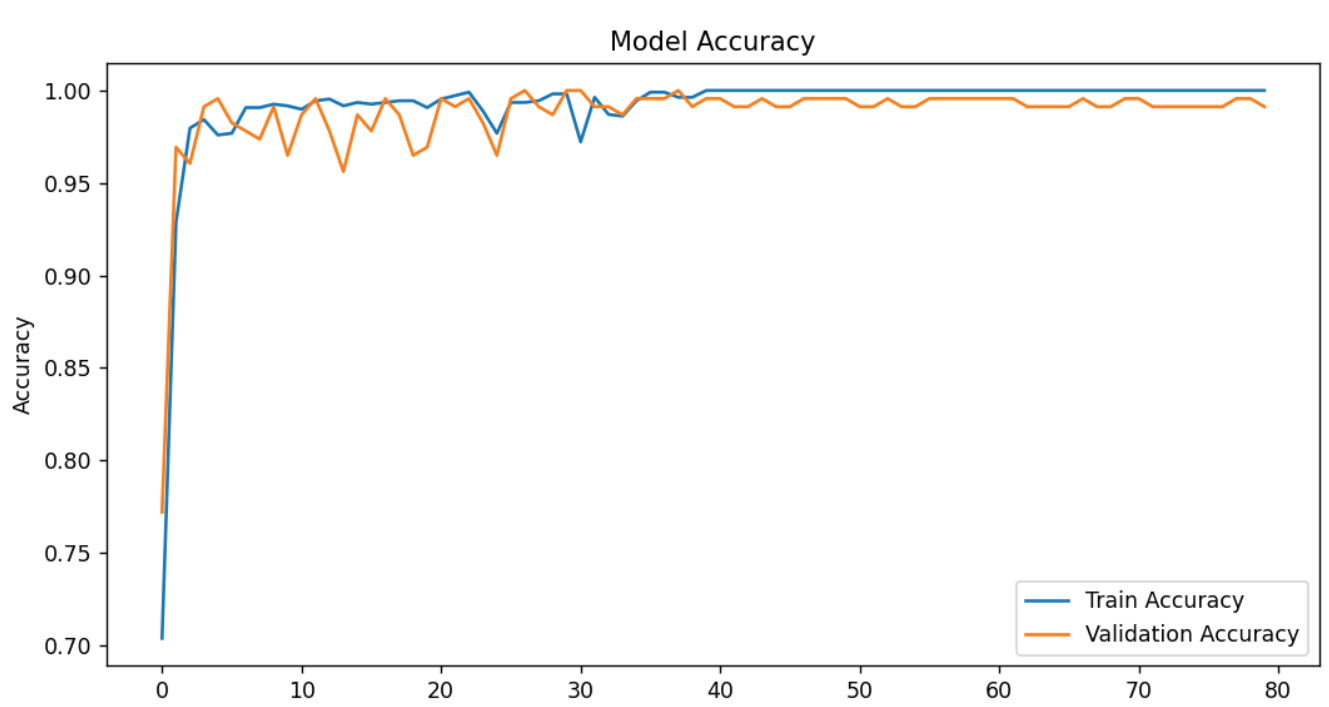

**Fig. 9. Accuracy over Epochs.** Source: The authors.

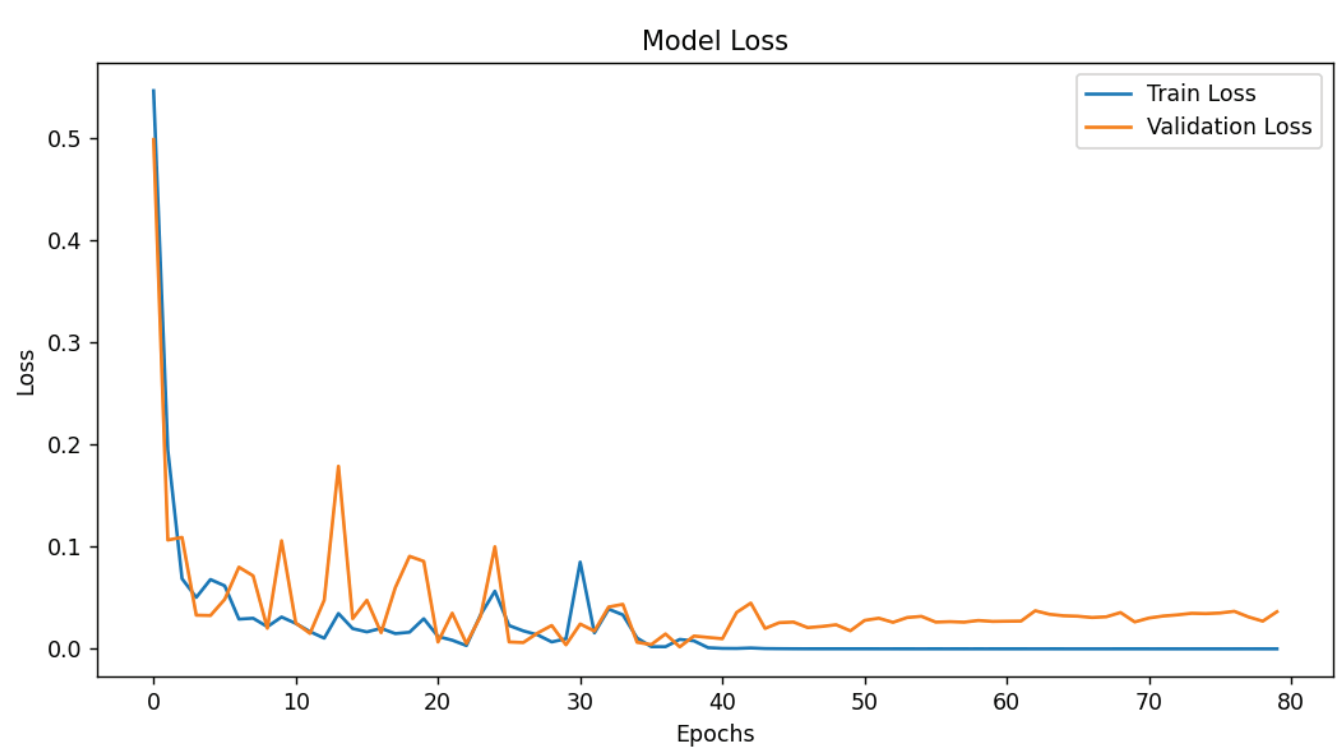

**Fig. 10. Error over Epochs.** Source: The authors.

The model parameters obtained after 80 iteration cycles on the test dataset are listed in Table 3. In validation experiments, the proposed system achieved over 99% accuracy on the test dataset.

**Table 3. Test model parameters.**

| Parameters | Percentage |
|---|---|
| Accuracy | 99.56 |
| Precision | 100 |
| Recall | 99 |
| F1-score | 100 |

Furthermore, the proposed system was applied in real-time experiments to observe various subjects (people) and provide instant warnings when they fall. The participants in the experiments included individuals aged between 8 and 50, both male and female. Each experiment lasted from 2 to 5 minutes and involved one participant within the camera field of view of the proposed fall detection system. The participant performed normal walking actions along with at least one unexpected fall action. A total of 20 fall actions were recorded throughout the real-time experiment. The experimental results detected 19 out of 20 fall actions, achieving a detection rate of 95%. Some illustrative images of fall detection are shown in Figs. 11, 12, 13. These images were extracted from real-time experiments conducted at the author's home. In these experiments, the proposed system achieved over 95% accuracy in real-time testing while also demonstrating good performance. From the real-time experiments, we conclude that the fall detection system operates efficiently, with a frame processing time ranging from 0.02 to 0.11 seconds and a response time consistently under 0.09 seconds, ensuring near-instantaneous reaction. This enables the system to respond quickly and accurately, making it suitable for real-time applications such as elderly care and security monitoring.

Finally, the research results are applied to elderly care and smart health monitoring environments. In elderly care, the system's non-invasive, camera-based design integrates seamlessly into smart homes, eliminating the discomfort associated with wearable devices and promoting user acceptance among seniors. Deployed on embedded devices (e.g., Jetson Nano, Raspberry Pi, etc.), it enables continuous monitoring in living spaces, delivering instant fall alerts to caregivers via IoT platforms (e.g., message queuing telemetry transport - MQTT), thereby reducing response times to less than one second. For smart health monitoring, integration with telehealth systems facilitates real-time data sharing with medical professionals, enhancing fall risk assessments and personalised care plans. The scalability to multi-camera setups can expand coverage in care facilities, mitigating occlusions. These applications enhance elderly safety, independence, and healthcare efficiency, with potential for broader adoption in smart healthcare ecosystems.

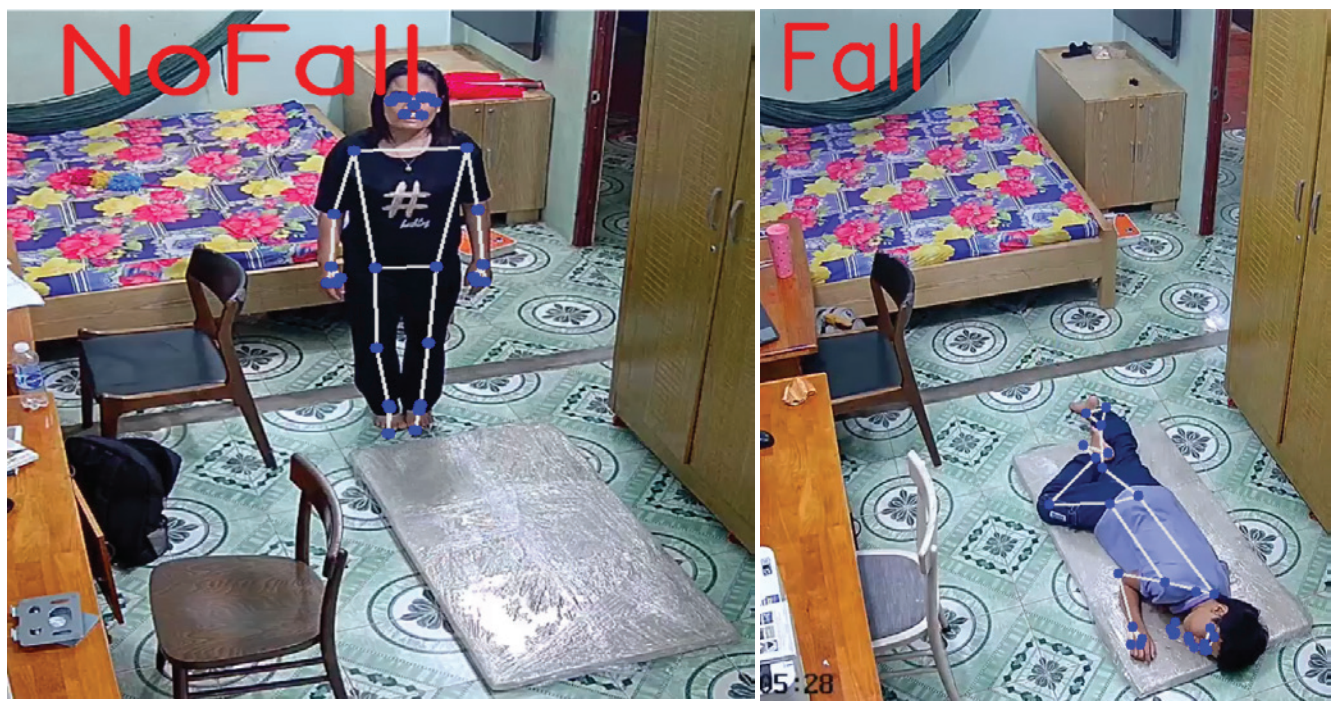

**Fig. 11. Images of fall detection 1.** Source: The authors.

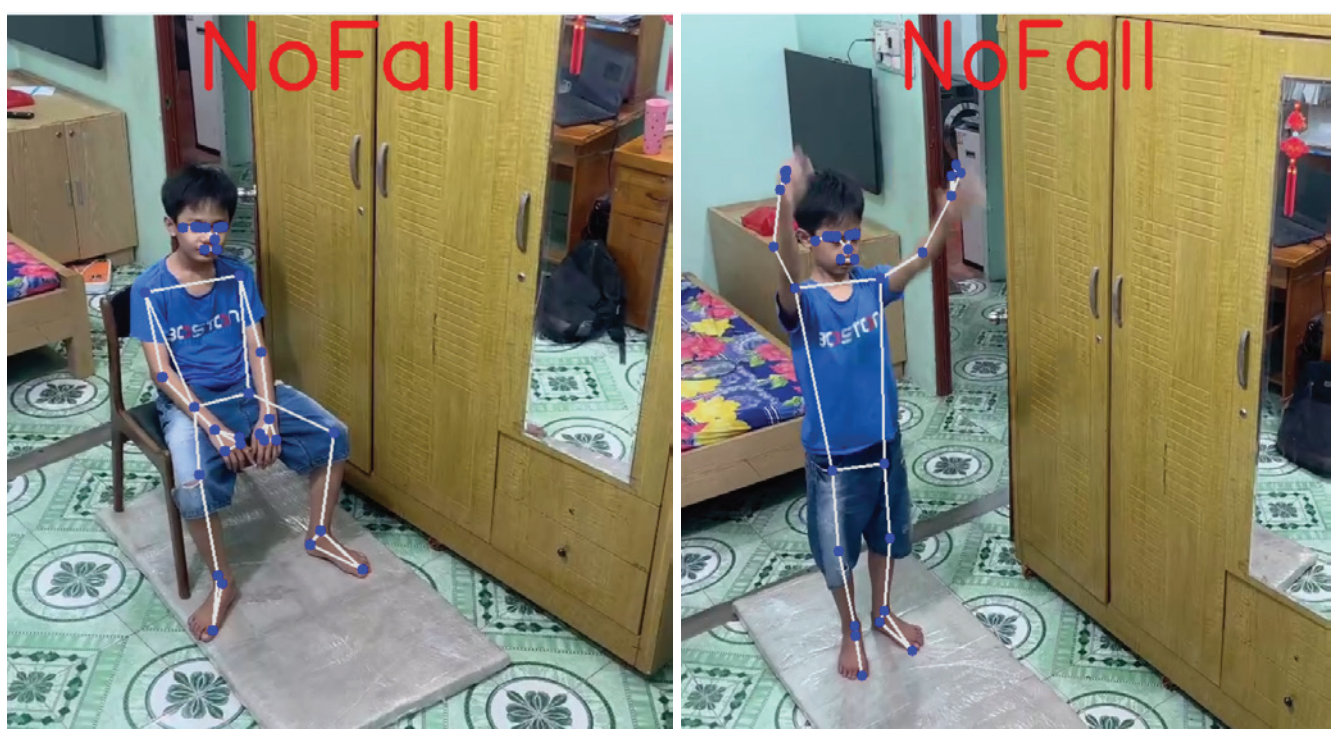

**Fig. 12. Images of no-fall detection.** Source: The authors.

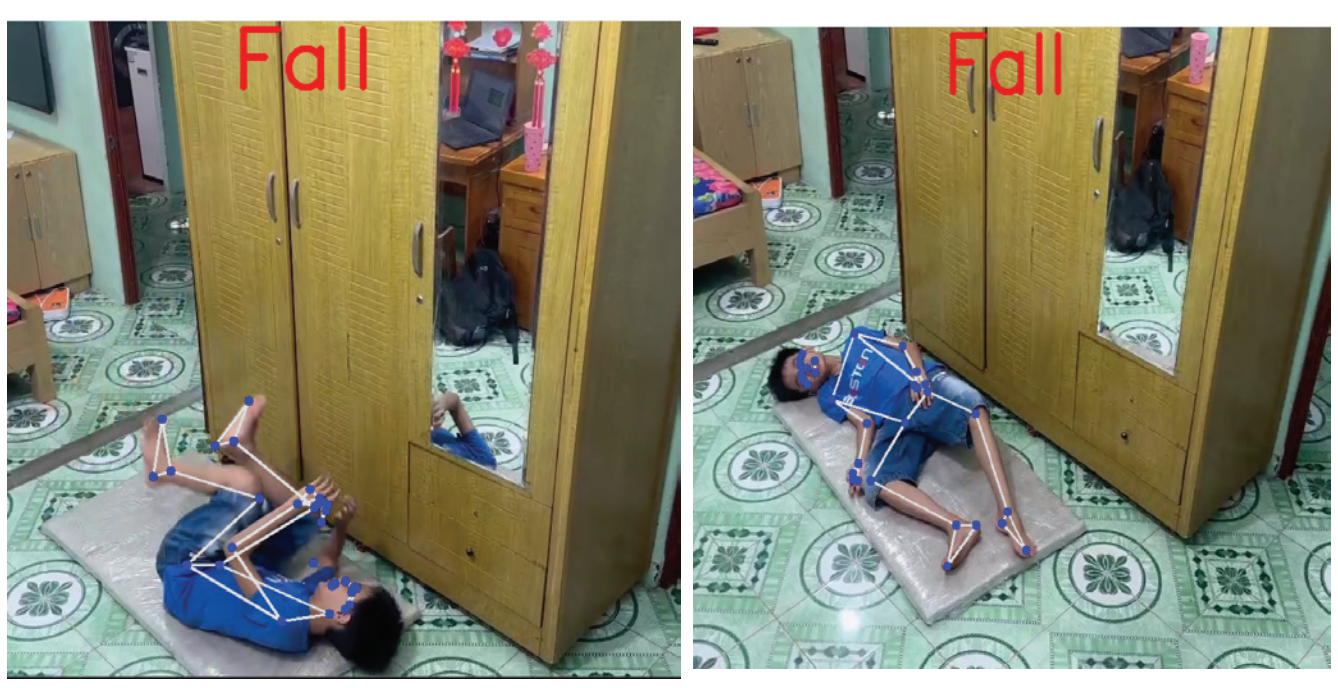

**Fig. 13. Images of fall detection 2.** Source: The authors.

## 5. Conclusions

The proposed vision-based fall detection system, leveraging MediaPipe's pose estimation and LSTM networks, achieves over 95% accuracy across diverse datasets, surpassing traditional wearable systems (80-90%) and ambient systems (85-90%). By extracting 33 body keypoints and modelling temporal motion patterns, it distinguishes falls from daily activities with a false-positive rate below 5%, compared to 15-20% in sensor-based methods. Real-time performance is robust, with frame processing times of 0.02-0.11 seconds and response times under 0.09 seconds, ensuring timely alerts for elderly care. Testing in varied environments, including living rooms, care facilities, and lighting conditions (50-1000 lux), yields F1-scores above 0.93, outperforming lab-constrained vision systems (0.85-0.90 F1). Its non-invasive, camera-based design enhances user acceptance, making it ideal for long-term smart home monitoring.

However, environmental conditions significantly impact performance. In low-light settings (<50 lux), keypoint detection accuracy drops by approximately 10% (e.g., from 95 to 85%), increasing missed detections by 8-12% due to reduced contrast and noise in RGB frames. High-lux variability (>1000 lux) causes overexposure, degrading accuracy by 5-7% in 10% of test cases. Occlusions from furniture or multiple occupants elevate false negatives by 3-5%, particularly in cluttered spaces (e.g., with <50% visible keypoints), as MediaPipe struggles with partial body tracking. Reliance on a single camera limits coverage in large areas (>20 m²), resulting in missed falls for 5-10% of incidents occurring outside the field of view. The computational demands of LSTM inference risk delays on low-power devices. Privacy concerns also arise, as continuous video capture necessitates secure data handling to prevent unauthorised access.

Future research should address these challenges. Integrating infrared or depth cameras could improve low-light accuracy to over 97%, mitigating 80% of missed detections. Multi-camera setups may reduce occlusions and expand coverage, boosting detection rates by 5-10%. Lightweight GRU or 1D-CNN models could reduce inference times by approximately 30%, enhancing embedded efficiency. IoT-enabled edge computing with MQTT protocols would enable sub-one-second caregiver alerts, improving response times by 50%. Developing privacy-preserving techniques, such as on-device processing or blockchain-inspired encryption, would safeguard data, aligning with smart home trends. These advancements would enhance the system's robustness, scalability, and ethical deployment, solidifying its impact on elderly safety and smart healthcare.

## CRediT author statement

Tri Nhut Do: Conceptualisation, Methodology, Formal analysis and investigation, Writing - Original draft preparation, Writing - Reviewing and Editing, Validation; Thi Thuy Le: Conceptualisation, Methodology, Writing - Reviewing and Editing, Validation, Resources, Supervision.

## ACKNOWLEDGEMENTS

This research was supported by the University of Information Technology, Vietnam National University, Ho Chi Minh City's Scientific Research Support Fund.

## COMPETING INTERESTS

The authors declare that there is no conflict of interest regarding the publication of this article.

## REFERENCES

[1] M.J.A. Nahian, J.F. Raisa, M. Mahmud, et al. (2026), "Artificial intelligence for elderly fall detection: State-of-the-art methods, applications and challenges", *Cogn. Comput.*, **18**, DOI: 10.1007/s12559-026-10550-5.

[2] N.T. Diep, D.A. Tuan, T.V. Nguyen, et al. (2025), "Associations of sleep quality and fall risk among older adult outpatients at Thai Binh Medical University Hospital, Northern Vietnam", *Front. in Sleep*, **3**, DOI: 10.3389/frsle.2024.1486794.

[3] L.T. Nguyen, K.G. To, T.C. Tang, et al. (2024), "Risk factors and profiles of falls among inpatients in Vietnam: A multicenter nested case-control study", *Risk Manag. Healthc. Policy*, **17**, pp.2229-2239, DOI: 10.2147/RMHP.S471895.

[4] K. Chouhan, A. Kumar, A.K. Chakraverti, et al. (2022), "Human fall detection analysis with image recognition using convolutional neural network approach", *Proceedings of Trends in Electronics and Health Informatics*, Springer, pp.95-106.

[5] C. Ge, I.Y.H. Gu, J. Yang (2018), "Co-saliency-enhanced deep recurrent convolutional networks for human fall detection in e-healthcare", *Annual International Conference of The IEEE Engineering in Medicine and Biology Society (EMBC)*, pp.1572-1575, DOI: 10.1109/EMBC.2018.8512586.

[6] T.N. Do, Y.S. Suh (2011), "Foot motion tracking using vision", *2011 IEEE 54th International Midwest Symposium on Circuits and Systems*, pp.1-4, DOI: 10.1109/MWSCAS.2011.6026603.

[7] T.B.N. Tat, T.T. Le, T.N. Do (2024), "IoT-based electrical energy monitoring system", *Future Data and Security Engineering. Big Data, Security and Privacy, Smart City and Industry 4.0 Applications*, Springer, DOI: 10.1007/978-981-96-0437-1_27.

[8] T.N. Do (2024), "IoT-based remote control for robotic arm", *International Journal of Systems, Control and Communications*, **15(2)**, pp.146-158, DOI: 10.1504/IJSCC.2024.138541.

[9] T.N. Do, C.L. Le, M.S. Nguyen (2021), "IoT-based security with facial recognition smart lock system", *15th International Conference on Advanced Computing and Applications*, pp.181-185, DOI: 10.1109/ACOMP53746.2021.00032.

[10] World Health Organisation (2021), *Falls*, https://www.who.int/news-room/fact-sheets/detail/falls, accessed 12 December 2021.

[11] R. Vaishya, A. Vaish (2020), "Falls in older adults are serious", *Indian J. Orthop.*, **54(1)**, pp.69-74, DOI: 10.1007/s43465-019-00037-x.

[12] D.J. Warrington, E.J. Shortis, P.J. Whittaker (2021), "Are wearable devices effective for preventing and detecting falls: An umbrella review (a review of systematic reviews)", *BMC Public Health*, **21**, DOI: 10.1186/s12889-021-12169-7.

[13] A.S.O. Bustos, M. Tramontano, G. Morone, et al. (2023), "Ambient assisted living systems for falls monitoring at home", *Expert Review of Medical Devices*, **20(10)**, pp.821-828, DOI: 10.1080/17434440.2023.2245320.

[14] A.E. Kaid, K. Baina, J. Baina (2019), "Reduce false positive alerts for elderly person fall video-detection algorithm by convolutional neural network model", *Procedia Computer Science*, **148**, pp.2-11.

[15] S. Negi, M. Garg, H. Maindola, et al. (2023), "Real-time human pose estimation: A MediaPipe and Python approach for 3D detection and classification", *2023 3rd International Conference on Technological Advancements in Computational Sciences*, pp.128-133, DOI: 10.1109/ICTACS59847.2023.10390506.

[16] P. Sirikongtham, A. Nimkoompai (2025), "A new method for real-time fall detection based on MediaPipe pose estimation and LSTM", *International Journal of Advanced Computer Science and Applications (IJACSA)*, **16(8)**, DOI: 10.14569/IJACSA.2025.0160811.

[17] C.A.Q. Bugarin, J.M.M. Lopez, S.G.M. Pineda, et al. (2022), "Machine vision-based fall detection system using MediaPipe Pose with IoT monitoring and alarm", *2022 IEEE 10th Region 10 Humanitarian Technology Conference*, pp.269-274, DOI: 10.1109/R10-HTC54060.2022.9929527.

[18] S. Saraswat, G. Malathi (2024), "Pose estimation based fall detection system using MediaPipe", *2024 10th International Conference on Communication and Signal Processing*, pp.1733-1738, DOI: 10.1109/ICCSP60870.2024.10543522.

[19] M.E. Karar, H.I. Shehata, O. Reyad (2022), "A survey of IoT-based fall detection for aiding elderly care: Sensors, methods, challenges and future trends", *Applied Sciences*, **12(7)**, DOI: 10.3390/app12073276.

[20] T. Alanazi, G. Muhammad (2022), "Human fall detection using 3D multi-stream convolutional neural networks with fusion", *Diagnostics*, **12(12)**, DOI: 10.3390/diagnostics12123060.

[21] R. Espinosa, H. Ponce, S. Gutiérrez, et al. (2020), "A vision-based approach for fall detection using multiple cameras and convolutional neural networks: A case study using the UP-fall detection dataset", *Computers in Biology and Medicine*, **115**, DOI: 10.1016/j.compbiomed.2019.103520.

[22] J. Zhang, J. Li, W. Wang (2021), "A class-imbalanced deep learning fall detection algorithm using wearable sensors", *Sensors*, **21(19)**, DOI: 10.3390/s21196511.

[23] W.L. Hsu, J.X. Liu, C.C. Yang, et al. (2023), "A fall detection system based on FMCW radar range-doppler image and Bi-LSTM deep learning", *IEEE Sensors Journal*, **23(18)**, pp.22031-22039, DOI: 10.1109/JSEN.2023.3300994.

[24] A. Ometov, V. Shubina, L. Klus, et al. (2021), "A survey on wearable technology: History, state-of-the-art and current challenges", *Computer Networks*, **193**, DOI: 10.1016/j.comnet.2021.108074.

[25] Q. Zhang, Y. Su, P. Yu (2014), "Assisting an elderly with early dementia using wireless sensors data in smarter safer home", *Service Science and Knowledge Innovation*, Springer, DOI: 10.1007/978-3-642-55355-4_41.